\documentclass[10pt,twocolumn,letterpaper]{article}
\usepackage[square,sort,comma,numbers]{natbib}
\usepackage{cvpr}
\usepackage{times}
\usepackage{epsfig}
\usepackage{graphicx}
\usepackage{amsmath}
\usepackage{amssymb}
\usepackage{natbib}
\usepackage{color}
\usepackage{subcaption}
\usepackage{float}

\usepackage[breaklinks=true,bookmarks=false]{hyperref}

\cvprfinalcopy 

\def\cvprPaperID{67} 

\begin{document}

\title{New Techniques for Preserving Global Structure and Denoising with Low Information Loss in Single-Image Super-Resolution}


\author{Yijie Bei \and \hspace*{-5pt} 
Alex Damian \and \hspace*{-5pt} 
Shijia Hu \and \hspace*{-5pt} 
Sachit Menon \and \hspace*{-5pt} 
Nikhil Ravi \and \hspace*{-5pt} 
Cynthia Rudin\thanks{All authors contributed equally. Thanks to other members of Duke Data Science Team} \and Duke University}


\maketitle

\begin{abstract}
  This work identifies and addresses two important technical challenges in single-image super-resolution: (1) how to upsample an image without magnifying noise and (2) how to preserve large scale structure when upsampling. We summarize the techniques we developed for our second place entry in Track 1 (Bicubic Downsampling), seventh place entry in Track 2 (Realistic Adverse Conditions), and seventh place entry in Track 3 (Realistic difficult) in the 2018 NTIRE Super-Resolution Challenge. Furthermore, we present new neural network architectures that specifically address the two challenges listed above: denoising and preservation of large-scale structure. 
\end{abstract}

\section{Introduction}
Super-resolution (SR) is a classic problem in image processing where the goal is to generate a high resolution image from one or more low resolution images. Applications of super-resolution are wide-ranging. For instance, SR is important for allowing modern high-definition displays to function properly when showing video recorded at lower resolutions. SR also has many applications in medical imaging, such as reducing noise in images stemming from uncontrollable patient motions \cite{Robinson1}. This work focuses on single image super-resolution, which is useful for photographic enhancement, license plate recognition, satellite imaging, and other remote sensing applications such as recognition of a military target \cite{Yue2}. 

Deep learning techniques can learn a mapping directly from low resolution to high resolution images, where all feature construction is automated. This makes some types of complex preprocessing much easier than previous approaches, for example, we no longer need to explicitly choose a dictionary of low-level features (e.g., edge detectors) to convolve with the image. The fact that training deep neural networks has become much easier within the past few years has led to more reliable automated training. On the other hand, the fact that these deep learning methods use recursive mathematical formulas that are now much more complicated than before makes it more difficult to determine how to best troubleshoot them to achieve higher-quality performance.

In this work we discuss several insights into the problem of single-image super-resolution -- many of which have led to higher quality performance beyond entries from last year's NTIRE single-image SR competition. These insights concern the amplification of noise when upsampling and the preservation of large scale structure in enhanced images. We introduce neural network architectures for both the denoising problem (DeNoising for Super-Resolution -- DNSR) and the problem of preserving large-scales structure (Automated Decomposition and Reconstruction for Super-Resolution -- ADRSR). Additionally we present a set of tricks that provided boosts in SR performance. 

For denoising while upsampling, we present the DNSR (and more basic DNISR) architecture that concatenates two networks, where the first network is for denoising and the second is a baseline method for SR. This leverages domain knowledge that the noise should not have been in the low-resolution image in the first place and thus we should not amplify it.
Training these concatenated networks led to improvements in performance in Track 2 (realistic mild adverse conditions) and Track 3 (realistic difficult) of the NTIRE SR 2018 challenge. 

Modern methods for SR have trouble preserving large scale structure. Even if the high resolution images look realistic in local patches, the global structure (such as stripes that reach across the full image) can have serious visible faults. We present an architecture for preserving structure at multiple scales. In our network, ADRSR, the original image is downsampled multiple times, convolutions are performed on each of the downsampled images, and combined to form the final high-resolution image. This allows a multiscale reconstruction of the image that includes information about the larger scales before modeling information at the smaller scales.


The architectures for denoising and preserving large-scale structure can be used with any network blocks used for SR; we used convolutional blocks from EDSR \cite{EDSR} within our implementations, but these can be changed to any other blocks. DNISR or DNSR combine any network for denoising with any network for SR. 

Most of the ideas discussed here were not implemented in time for the NTIRE 2018 SR competition deadline. However, we present a set of tricks that were helpful in achieving higher level performance during the competition. For instance, an idea used in our Track 1 (classic bicubic downsampling) entry was to randomly shuffle the red, green, and blue layers of the image during training, which helps as a form of self-ensembling. We also discuss different upsampling techniques, and find that for x8 amplification, we should learn the fully amplified image directly, because learning a x4 followed by a x2 amplification tend to lead to the spurious addition of details that do not exist in the original high-resolution image. 
 
All of these ideas were developed over the course of approximately 8 weeks by a team of 5 undergraduates with no previous experience in image processing.

Our entries in the 2018 NTIRE superresolution competition \cite{Timofte_2018_CVPR_Workshops} achieved seventh place in Track 2 (realistic mild adverse conditions), seventh place in Track 3 (realistic difficult) and second place in Track 1 (classic bicubic downsampling). 
\begin{table}[ht]
    \centering
    \begin{tabular}{c|c|c|c}
         & Track 1 & Track 2 & Track 3\\ \hline
         PSNR & 25.433 & 23.374 & 21.658\\
         SSIM & 0.7067 & 0.6252 & 0.5400
    \end{tabular}
    \caption{Competition Result}
    \label{tab:my_label}
\end{table}

\section{Previous Work}
Many approaches to single-image super-resolution are based on different methods of image upsampling. In particular, nearest-neighbors upsampling (in which each unknown pixel in the upsampled image is assigned the value of its nearest known neighbor) and bicubic upsampling (in which each unknown pixel in the upsampled image is assigned a value interpolated from its nearest known neighbors) are popular methods for basic upsampling \cite{babu2011survey,gilman2006near}. These methods, while simple and computationally efficient, do not provide realistic high-resolution images. More advanced methods attempt to build a map between low resolution images and high resolution images through a variety of different techniques. Some techniques include frequency-domain methods such as alias removal \cite{tsai1984multiframe}, recursive least squares \cite{kim1990recursive}, and multichannel sampling theorem methods \cite{ur1992improved}, as well as spatial-domain methods, such as iterated back-projection \cite{irani1993motion}, joint MAP restoration \cite{hardie1997joint}, and adaptive filtering \cite{patti1998new}.

Neural networks have recently been successful for image processing tasks, and through application of classical ResNet architectures, Ledig et al$.$  created one successful example of a convolutional neural network for super-resolution, called SRResNet \cite{LEDIG}. Their work showed that the use of residual blocks improved performance on super-resolution tasks over more traditional convolutional neural network architectures, and has become the basis for many future architectures for super-resolution. Lim et al$.$ then improved on this with their EDSR method by removing batch normalization, using an L1 rather than L2 loss function, and adding depth to the network \cite{EDSR}. While these models have seen some success in the super-resolution task for `clean' images (that is, images that have been bicubically downscaled with no further degradations), they do not show good results for images with noise, blur, or other degradations. 

A few recent interesting super-resolution techniques have been suggested for degraded images. Zhang et al$.$ \cite{IRCNN} suggested using CNN denoisers as a modular part of model-based optimization methods to perform various computer vision tasks including super resolution. Shocher et al$.$ \cite{zssr} proposed an unsupervised approach that trains an image-specific CNN at test time that learns to use the repetitive structure of images to fill in details where there previously were none.

Other neural-network based methods, such as generative adversarial networks \cite{LEDIG}, have shown success in super-resolution as measured by human viewers. However, these networks achieve visual effects suitable for human viewing by `hallucinating' features from the low resolution image that are not necessarily in the original image, but would be believable given the low resolution image. As such, they are not as well suited for tasks that maximize similarity to the original high resolution image, such as PSNR and SSIM.

The methods introduced into this work are different in that they heavily leverage prior knowledge: DNSR leverages the knowledge that denoising before upsampling is helpful, while ADRSR uses a pyramid of downsampled images to borrow information at broader scales. The ideas within ADRSR and DNSR can be combined with any neural network approaches to denoising and super-resolution in order to include domain knowledge.



\section{Challenges}
When approaching all three super-resolution tracks (corresponding to non-noisy and noisy images), we encountered multiple challenges.

\begin{figure}[ht!]
    \centering
    \includegraphics[width=\columnwidth]{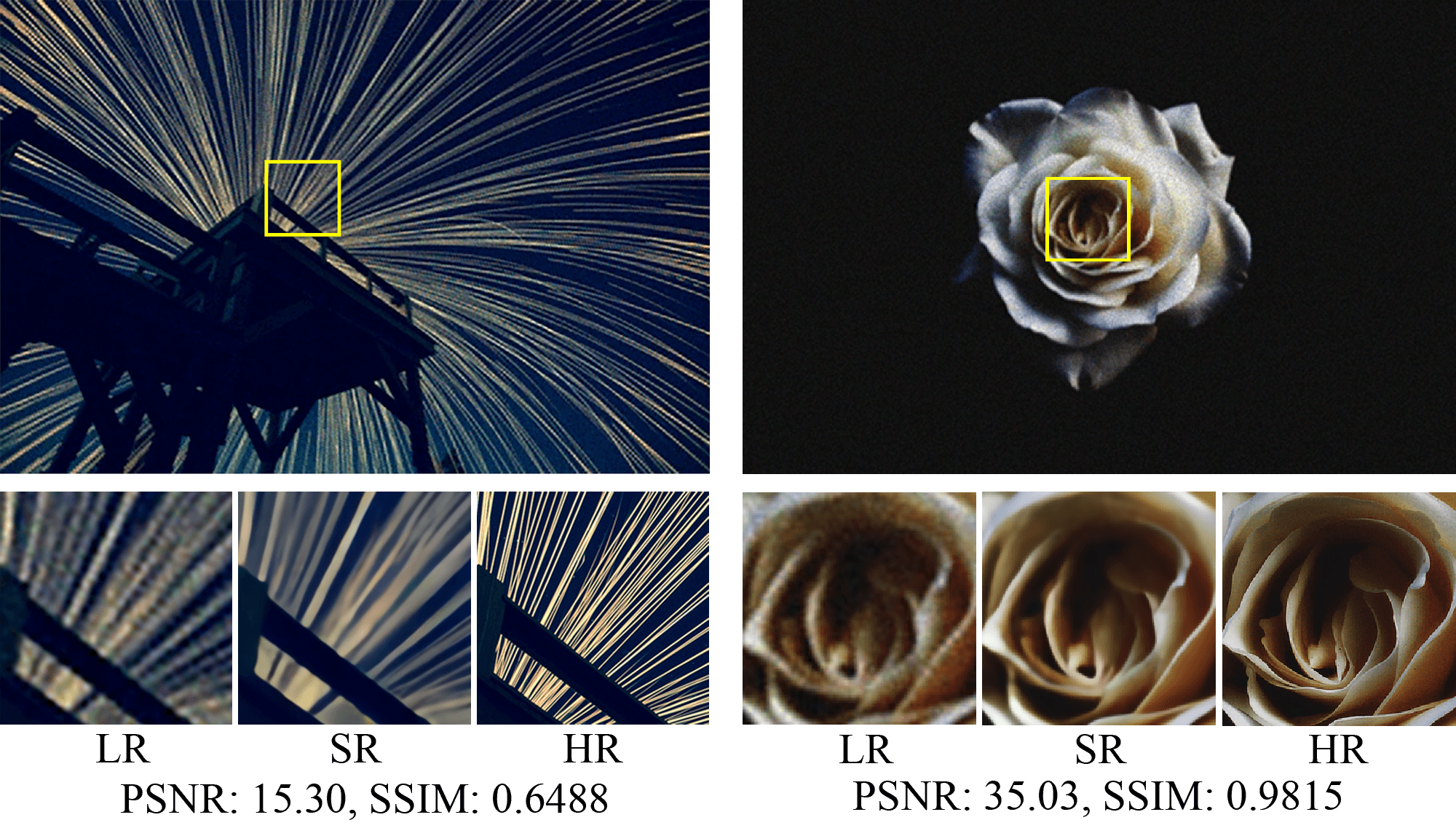}
    \caption{Varying PSNR between our algorithm's images for Track 2}
    \label{fig:varyingpsnr}
\end{figure}

First, there were challenges that were specific to the competition itself. One such challenge was that of \textit{model validation}, because the PSNR values of our algorithm varied wildly between images (see Figure \ref{fig:varyingpsnr}). Depending on which 100-image subset we used for validation, average PSNR values ranged from 22 to 27. This made it difficult to compare our results to others' and required us to fix a validation set of 100 images throughout training.

Particularly for noisy images, it is very difficult to avoid \textit{amplifying the noise while upsampling}. Several of the techniques we introduce here were useful for this, particularly the denoising and upsampling network DNSR for Tracks 2 and 3. Even without noise, artifacts tend to appear when upsampling by a factor of eight.



\begin{figure}
    \centering
    \includegraphics[width=\columnwidth]{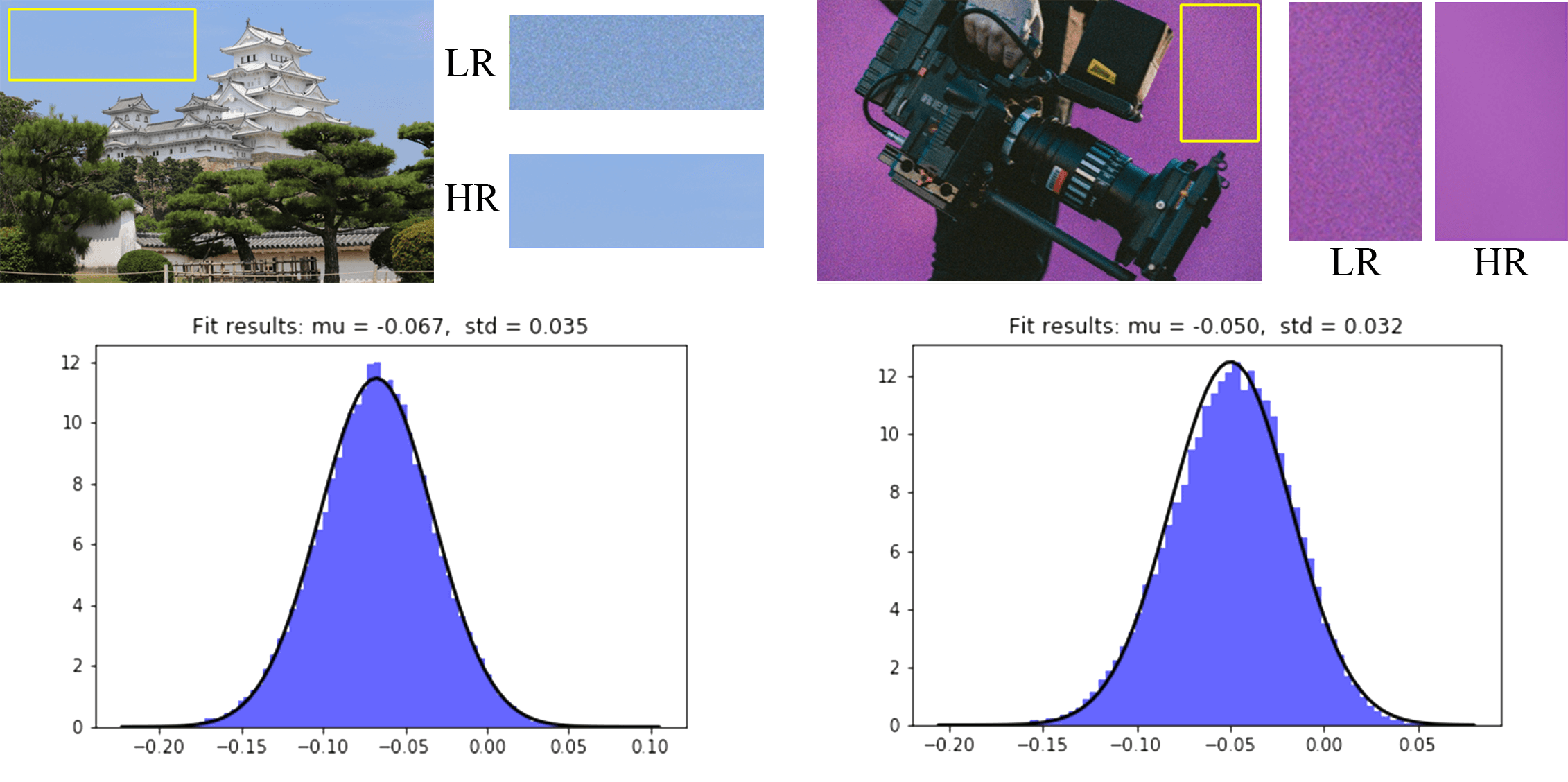}
    \caption{Histogram of noise from two images}
    \label{fig:hist}
\end{figure}

Most traditional denoisers require some knowledge of the noise itself, normally the standard deviation. To use any of these denoisers, it was imperative to \textit{reverse engineer} the noise. We took approximately flat areas of various images and considered the difference between the degraded low resolution images and down-scaled versions of the high resolution images. Because a blur kernel has no effect on flat regions of an image, this difference should be a good approximation of the noise (see Figure \ref{fig:hist}).


Most prior convolutional networks for super-resolution tend to focus on increasing the resolution in local areas; however, this approach does not \textit{take into account more global patterns} (such as zebra stripes). Some recent work \cite{upanddown,zssr} have aimed to solve this problem in other promising ways, and we present a new method for handling this (ADRSR) in what follows.



\section{ADRSR: A type of architecture that preserves global structure}

Figure \ref{fig:antialias} shows the types of problems that can arise from EDSR and similar SR algorithms. These algorithms consider local image patches, and do not aim to reconcile them with larger-scale patterns that crosscut into different patches.
Both increasing the depth of the network and increasing the size of each kernel allows the network to include larger scale patterns. However, these approaches are either hard to train, or do not converge at all. Thus, we reasoned that these larger patterns could be detected even by using a smaller kernel on a downsampled image without significant loss of information; the flexibility afforded by a large number of larger kernels may be unnecessary to capture this information.

\begin{figure*}[ht!]
\centering
\includegraphics[width=0.70\textwidth]{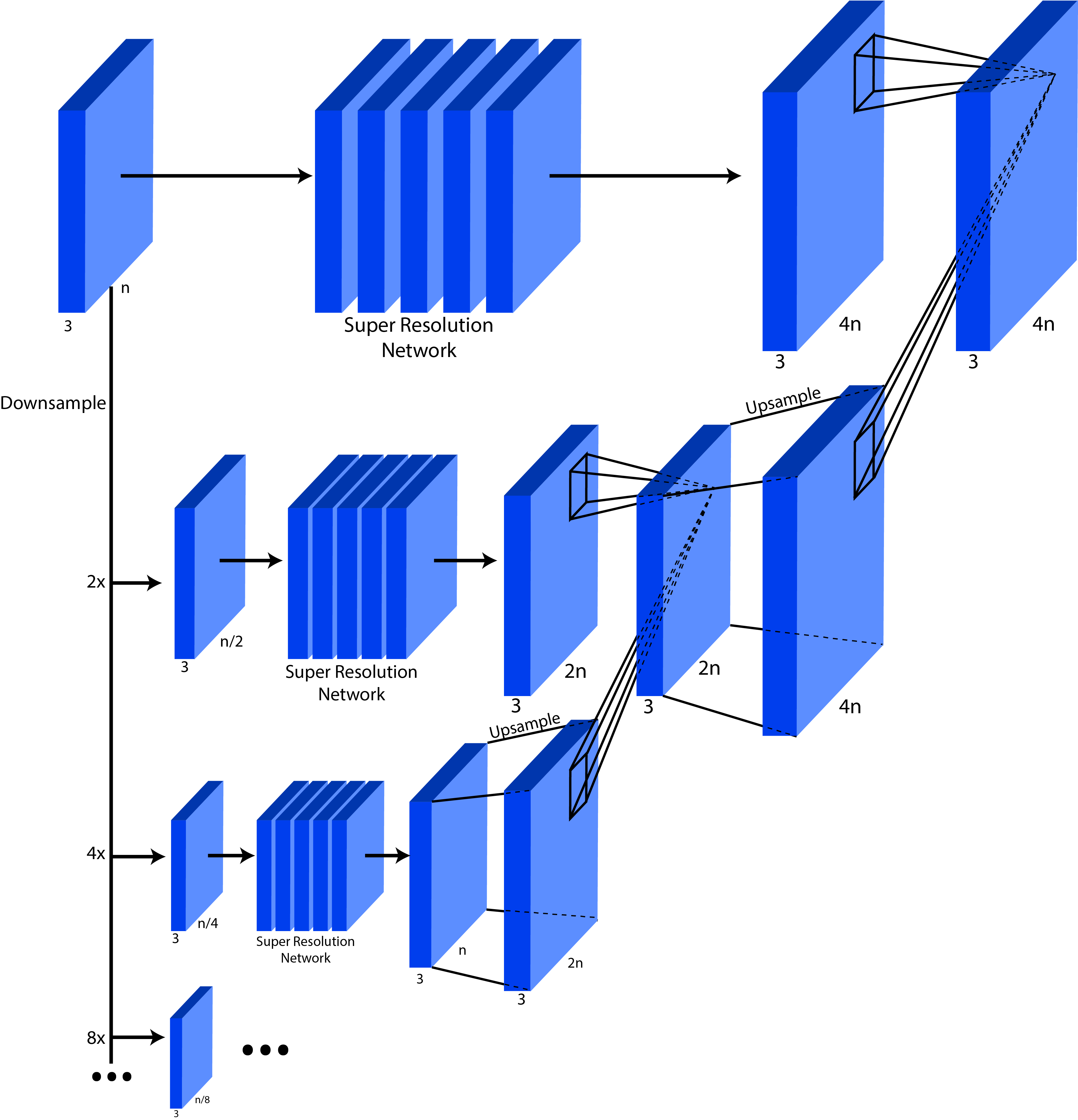}
\caption{ADRSR Network with a x4 super resolution network
\label{fig:multscalenet}}
\end{figure*}

The architecture that we introduce for preserving global structure is presented in Figure \ref{fig:multscalenet}, called Automated Decomposition and Reconstruction for SR (ADRSR). The original image is downsampled several times, with each downsampled image being fed through a parallel super-resolution network. This pyramid representation for the input allows us to create filters that capture patterns from the original image at various scales. We then iteratively combine the information from the various upscaled images to produce a final, more accurate image that respects global structure. When running the network forward on a new image, it would start from the coarsest scale, and iteratively add more detail on the finer scales. 

In Figure \ref{fig:multscalenet}, the SR network labeled in the figure can be replaced with any SR network. 

\begin{figure}[htbp]
	    \centering
	    \includegraphics[width=\columnwidth]{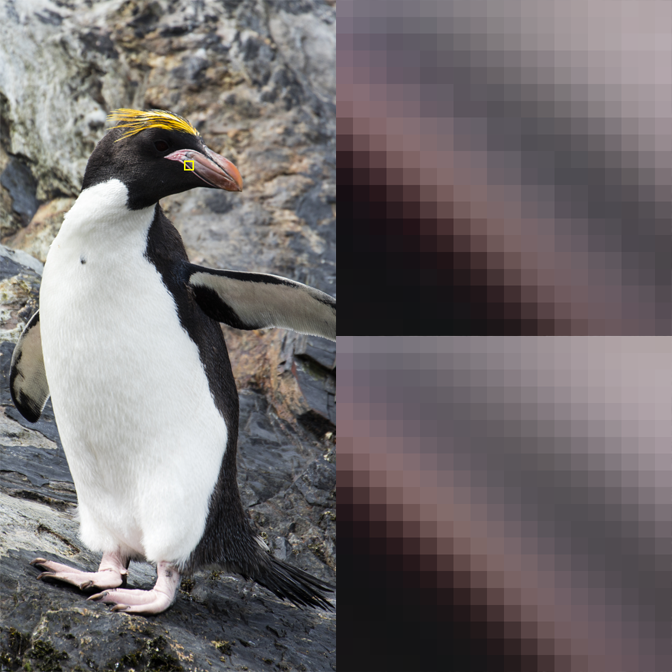}
	    \caption{Before unfreezing all of the weights, ADRSR tended to produce blocky artifacts (top-right), however after unfreezing all of the weights and training for a few more epochs, the artifacts disappeared (bottom-right).}
	    \label{fig:blockyup}
\end{figure}

\begin{figure*}[ht!]
    \centering
    
    \includegraphics[width=\textwidth]{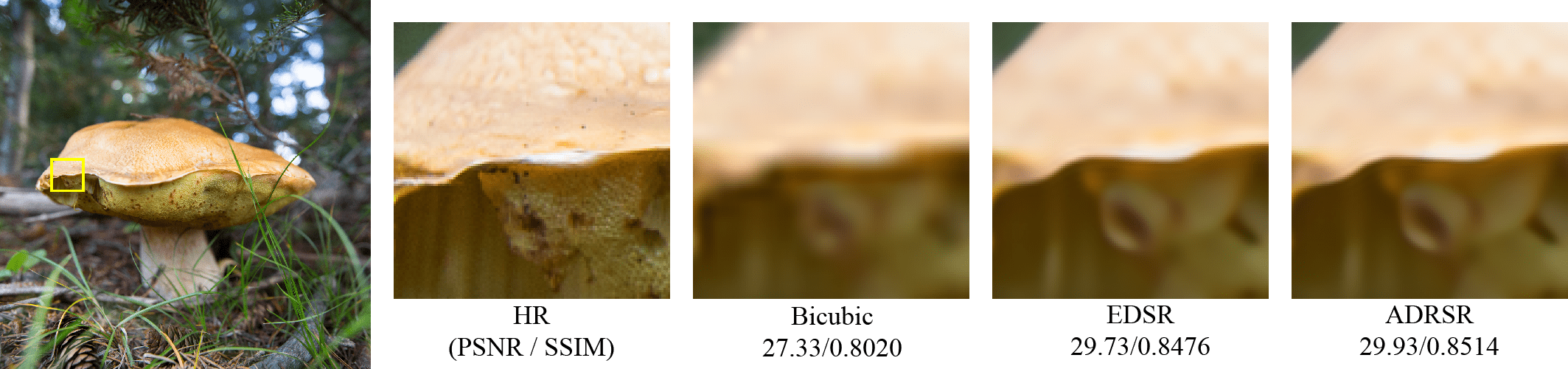}\\
    \includegraphics[width=\textwidth]{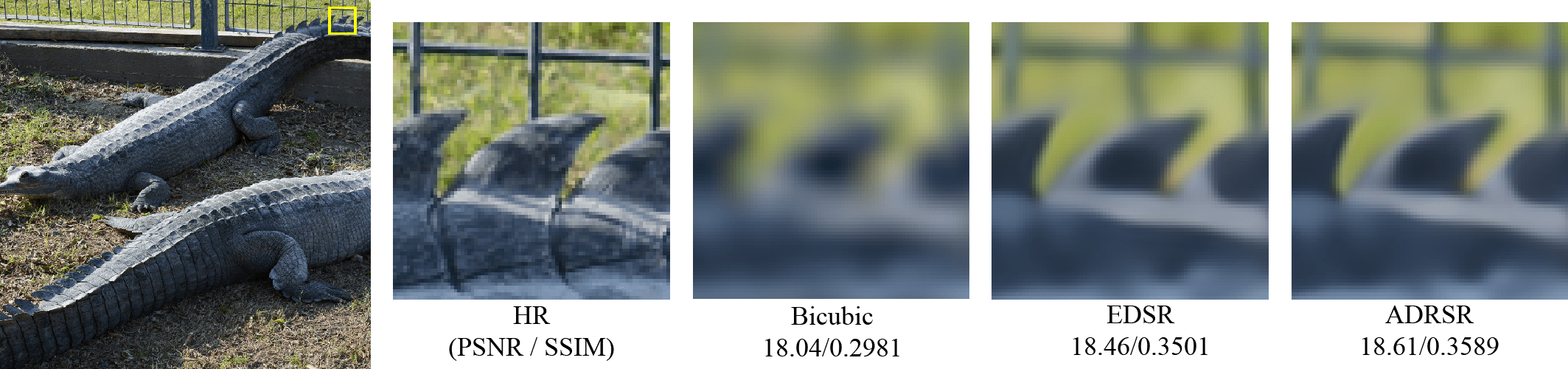}\\
    \includegraphics[width=\textwidth]{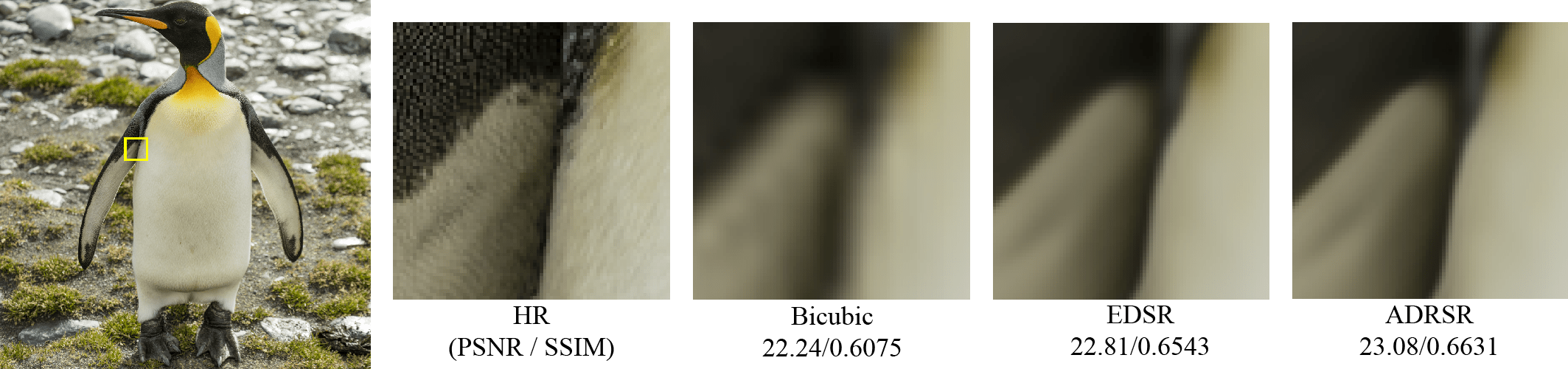}
    
    \caption{Comparison of baseline EDSR with ADRSR. Although the PSNR values of EDSR and ADRSR are similar, ADRSR tends to produce sharper lines and edges throughout the validation set.}
    \label{fig:adrsr}
\end{figure*}

We trained ADRSR to perform x8 upscaling using a baseline EDSR upscaler as the modular super resolution network. We initialized the bottom copy of EDSR with our fully trained baseline model, and then iteratively trained each successive level by temporarily removing all levels above it, and directly outputting the final result of that level (see Figure \ref{fig:multscalenet}). While training a level, we froze all weights except for those in the super resolution network in that level and the weights of the convolutional layer that combines the results of the current level with the result from the previous level. After training each level to convergence, we unfroze all of the weights and trained the entire network at once, which eliminated some blocky artifacts that appeared as a result of the upscaling process (see Figure \ref{fig:blockyup}). The results of this can be seen in Table \ref{tbl:adrsr} and Figure \ref{fig:adrsr}.

\begin{table}[ht!]
    \centering
    \begin{tabular}{c|c|c}
       \footnotesize{Algorithm}  & \footnotesize{EDSR} & \footnotesize{ADRSR}\\ \hline
        \footnotesize{PNSR} & \textcolor{red}{25.49} & 25.38\\
        \footnotesize{SSIM} & \textcolor{red}{0.6930} & 0.6898\\
    \end{tabular}
    \caption{Comparison of ADRSR to baseline EDSR
    \label{tbl:adrsr}}
\end{table}

While the results do not show any difference from EDSR in terms of numerical performance metrics, the network's multiscale reconstruction was intuitive and interesting. It is possible that this architecture could be useful for other applications besides PSNR/SSIM optimization.


\section{DNSR: A type of architecture for denoising with low information loss for SR}
\begin{figure}[htbp]
\centering
\includegraphics[width=\columnwidth]{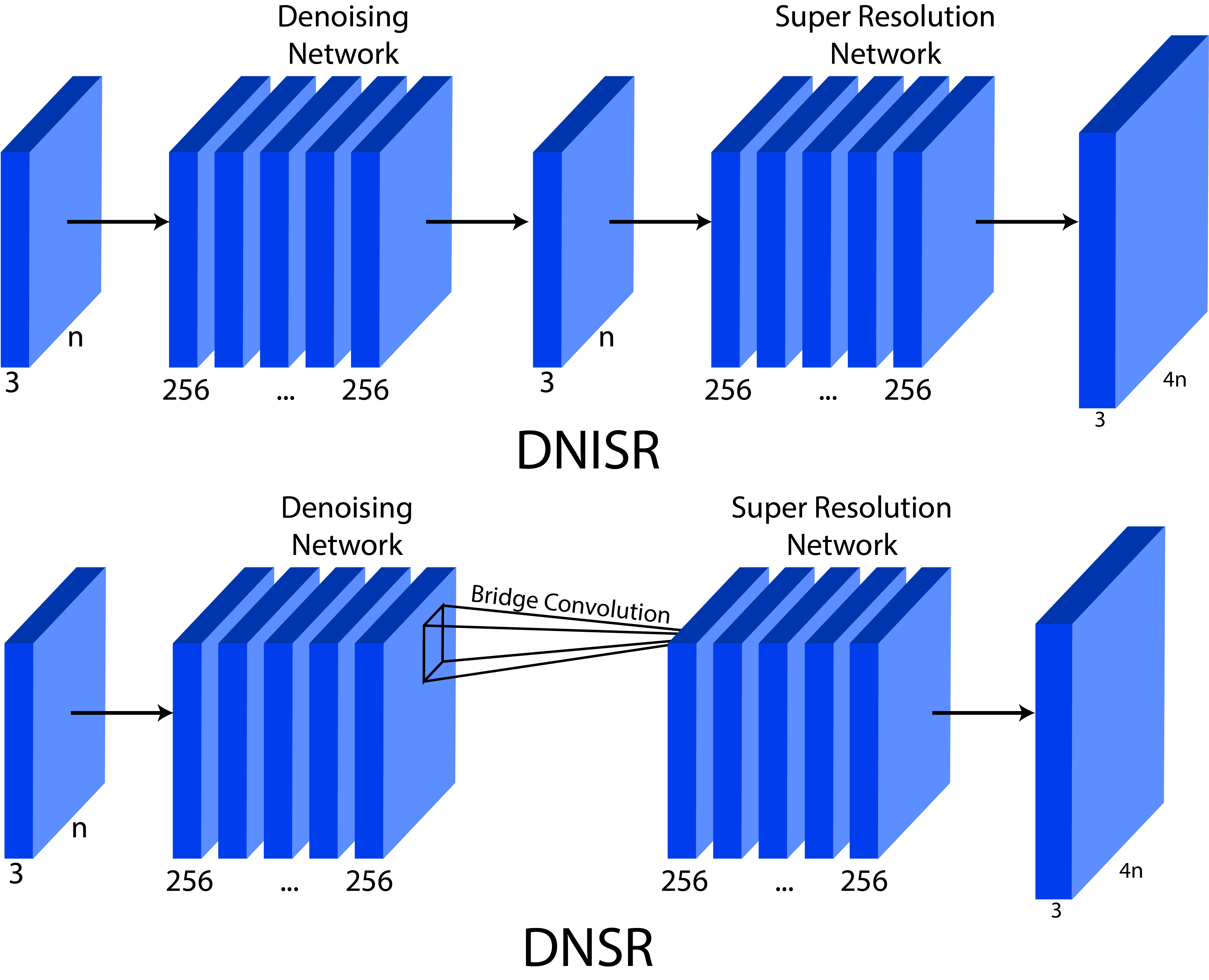}\\
\caption{Architectures for DNSR and DNISR.}
\label{fig:dnsr}
\end{figure}

As the principal challenge for Tracks 2 and 3 is noise, we considered three possible approaches for dealing with the noise:
\begin{itemize}
\item (Baseline simple approach). The simplest approach is to manually preprocess the images with a noise reduction algorithm, and then train a super resolution convolutional network on the denoised images.
\item (SR without denoising). Allow the residual blocks in a super-resolution network (such as EDSR) to simultaneously denoise the input images and extract features. That is, we directly train EDSR on the noisy data.
\item (DNISR, DNSR) After training the denoising network and SR network separately, we concatenate them, and then continue to train them together as a single network. DNISR and DNSR differ in the way that they concatenate the two networks during the final training stage, see Figure \ref{fig:dnsr}.
\end{itemize}

The first baseline approach allowed us to incorporate domain knowledge about the noise, but performed poorly due to the information loss caused by the denoiser. The second approach, on the other hand, did not tend to suffer from information loss. However, it was not possible to incorporate any domain knowledge about the problem (for instance that the image needs to be denoised) into the network. The third and fourth approaches solved both problems. They allowed us to incorporate domain knowledge into the network, since we could explicitly train the denoising network. DNSR trains the denoiser and super-resolution network together at the end to minimize information loss of the overall procedure. 
This approach is also advantageous when given a small number of images with the same degradations applied. After reverse-engineering the noise, external data can be used to train the denoising and super-resolution networks, and then the entire concatenated network can be trained on the dataset to allow the network to correct any additional degradations.

Based on our final approach, we constructed two models, which perform the concatenation in two different ways. The first was DNISR (DeNoising Into Super-Resolution), which ran the image through a denoising network (we used DNCNN \cite{DNCNN}), producing a low-resolution noise-reduced image, and then ran the result through the super-resolution network (we used EDSR) to produce a high-resolution image. 

We found a useful trick to further minimize information loss in DNISR: we fed the original image into the super-resolution network alongside the noise-free image with weights initialized to $0$. 

The second approach (DeNoising and Super-Resolution -- DNSR) used a more complicated concatenation procedure. It removed the information bottleneck between the two networks by combining the tail layer of the denoiser (which mapped $256$ channels to $3$) and the head layer of the SR network (which mapped $3$ channels to $256$) into a single bridge convolutional layer that mapped directly from the number of feature maps in the denoiser to the number of feature maps in the SR network. Unlike DNISR, there is no denoised image produced before entering the super-resolution network. See Figure \ref{fig:dnsr} for the architecture. We submitted the same model for Tracks 2 and 3 of the NTIRE 2018 competition.

\begin{figure}[htbp]
\centering
\includegraphics[width=\columnwidth]{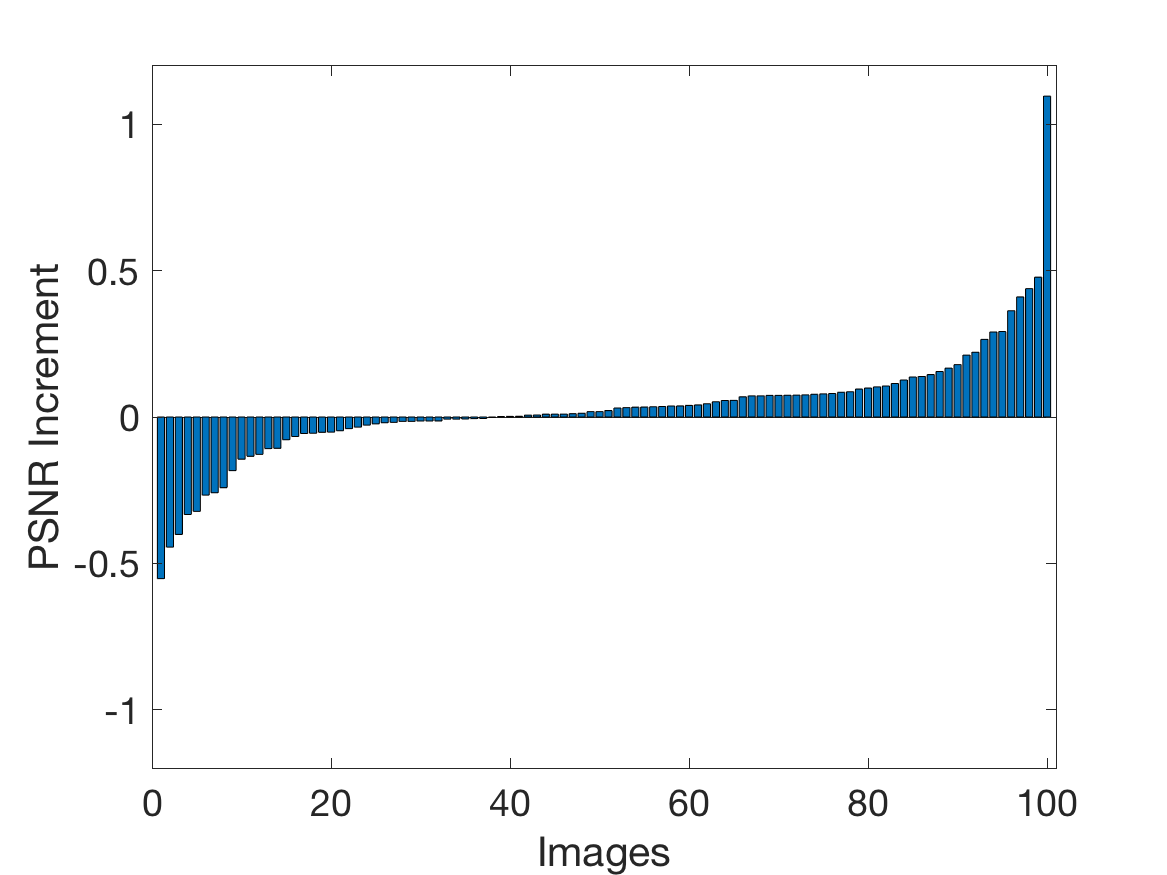}\\
\caption{PSNR difference between DNISR and EDSR (sorted by difference in PSNR) on the 100 image validation set from DIV2K \cite{Agustsson_2017_CVPR_Workshops}.}
\label{fig:track2diff2}
\end{figure}

\begin{figure}[htbp]
\centering
\includegraphics[width=\columnwidth]{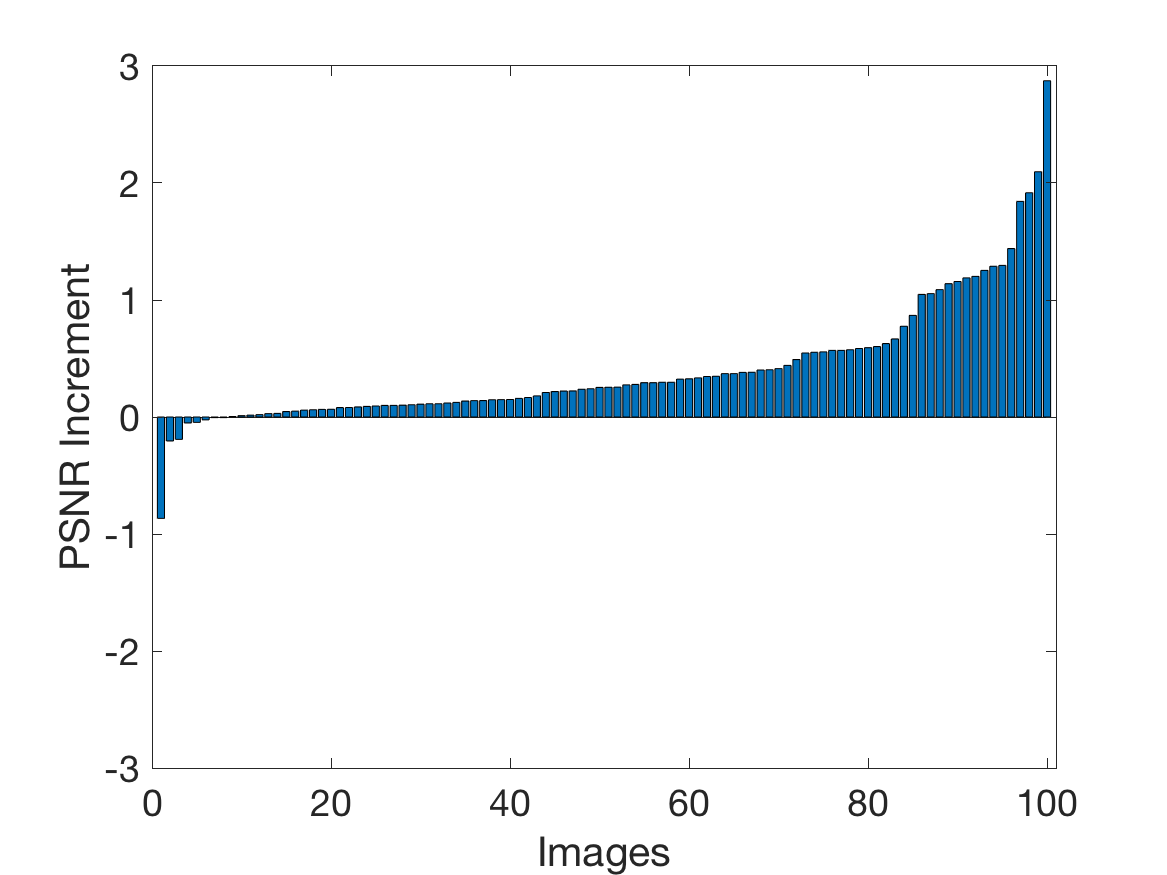}\\
\caption{PSNR difference between DNSR and EDSR (sorted by difference in PSNR) on the 100 image validation set from DIV2K \cite{Agustsson_2017_CVPR_Workshops}}
\label{fig:track2diff}
\end{figure}

Table \ref{tbl:dnsr} and Figures \ref{fig:track2diff2} and \ref{fig:track2diff} show a PSNR comparison for EDSR, DNISR, and DNSR. For a visual comparison of the images produced by each algorithm, see Figure \ref{fig:samples}.
\begin{table}[htbp]
    \centering
    \begin{tabular}{c|c|c|c|c}
       Algorithm & BICUBIC & EDSR & DNISR & DNSR \\\hline
       PNSR & 23.47 & 24.49 & 24.52 & \textcolor{red}{24.90} \\
       SSIM & 0.7333 & 0.7925 & 0.7940& \textcolor{red}{0.7956}
    \end{tabular}
    \caption{Comparison of results from EDSR and our denoising networks. The numbers reported were computed on the DIV2K \cite{Agustsson_2017_CVPR_Workshops} validation data set.}
    \label{tbl:dnsr}
\end{table}

\section{General Tricks and Insights}
\begin{figure}[ht]
    \centering
    \includegraphics[width=\columnwidth]{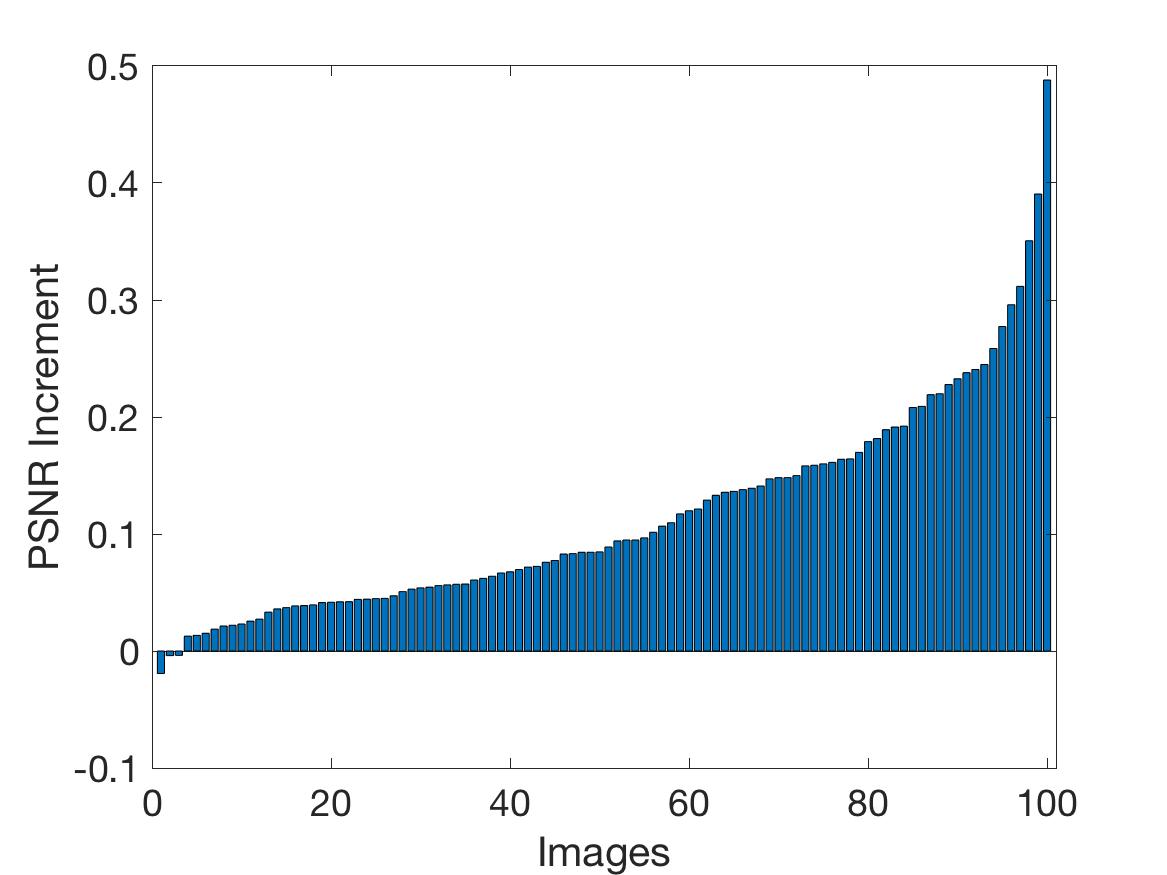}
    \caption{PSNR increment from test time RGB Shuffling (sorted by difference in PSNR)}
    \label{fig:rgbshuffling}
\end{figure}

\begin{figure*}[ht!]
\centering
\begin{subfigure}[t]{0.4\textwidth}
    \includegraphics[width=\textwidth]{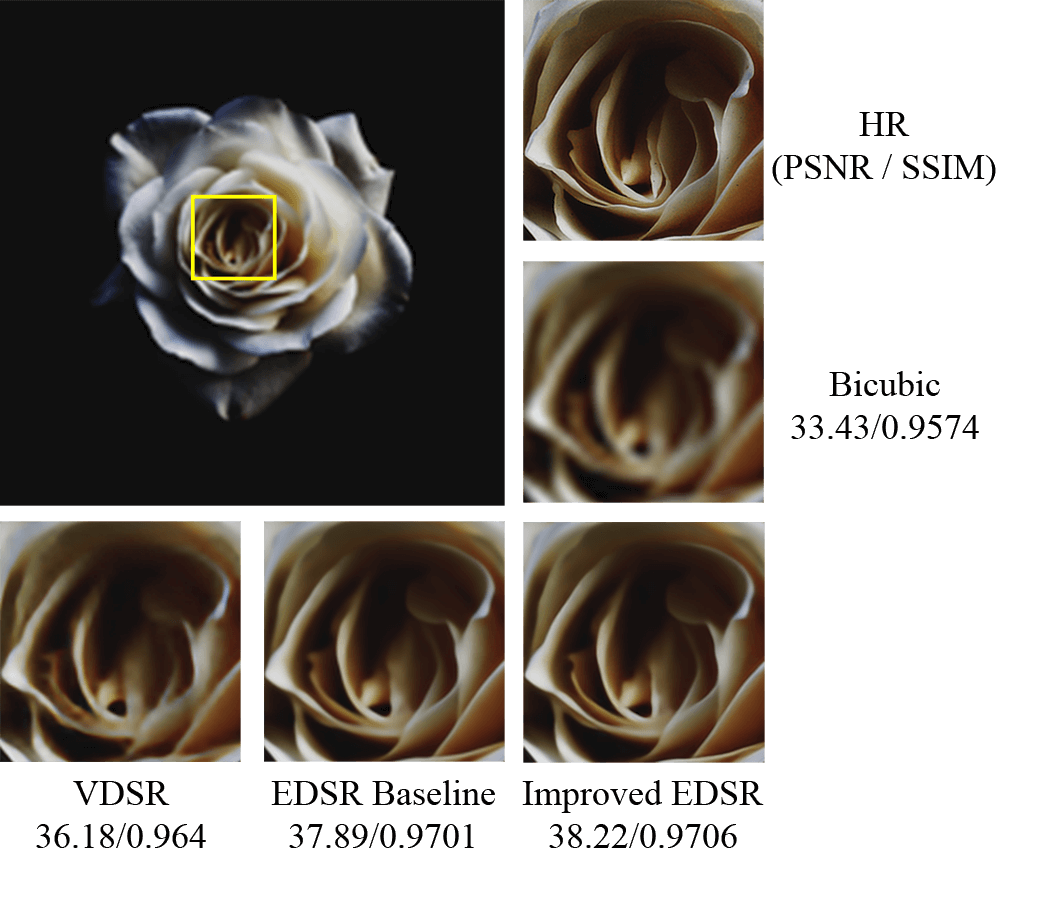}\\
    \includegraphics[width=\textwidth]{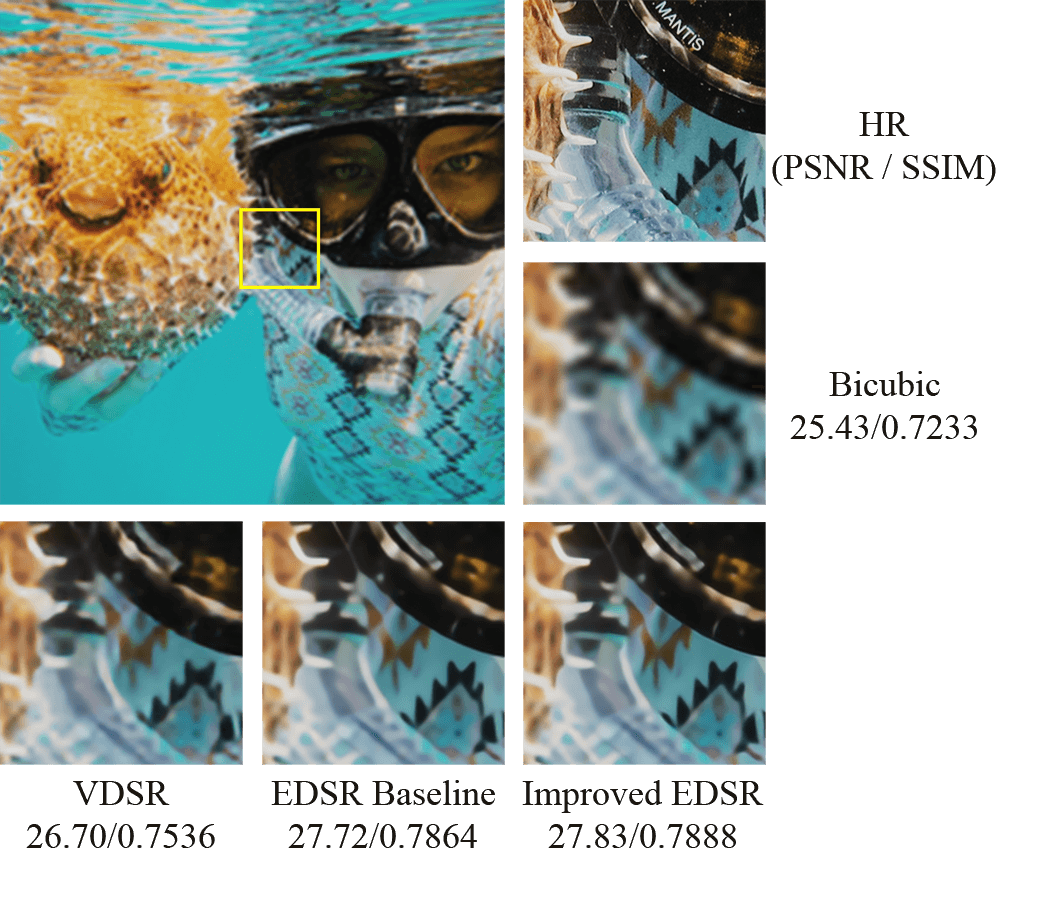}\\
    \includegraphics[width=\textwidth]{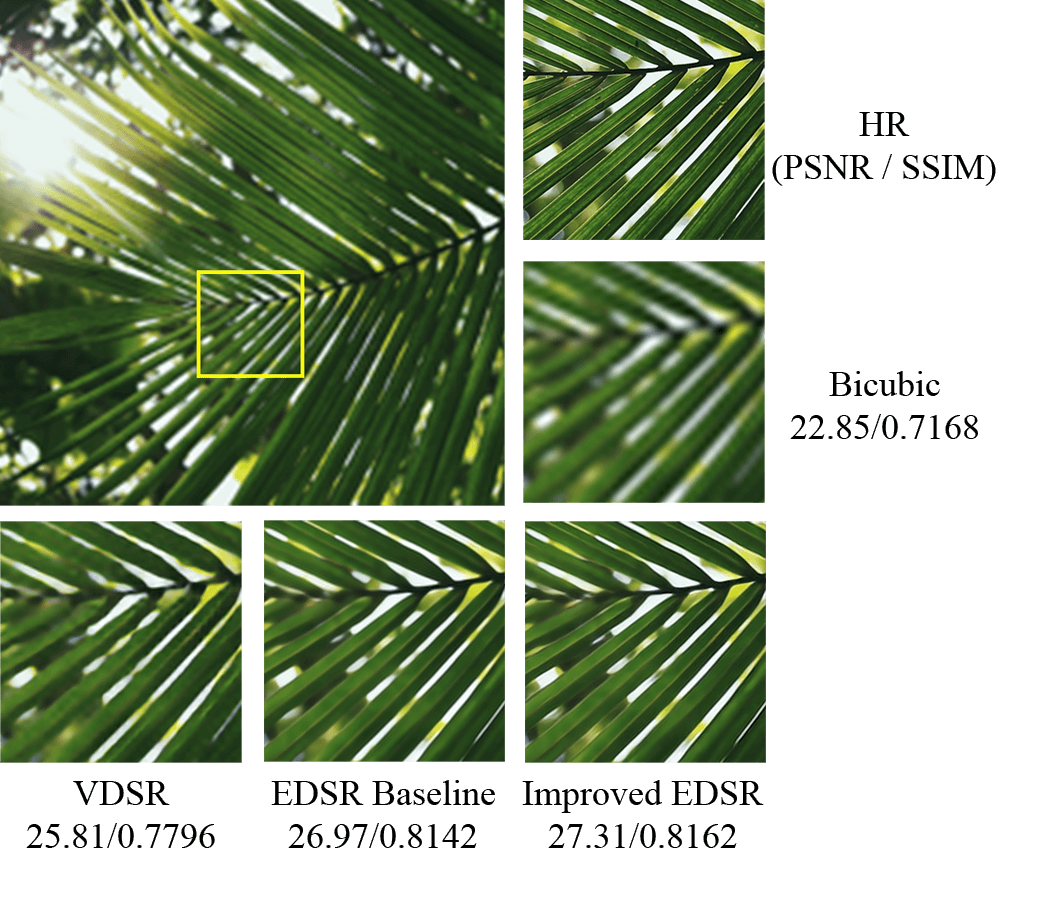}
\end{subfigure}
\hspace{0.05\textwidth}
\vrule
\hspace{0.1\textwidth}
\begin{subfigure}[t]{0.4\textwidth}
    \includegraphics[width=\textwidth]{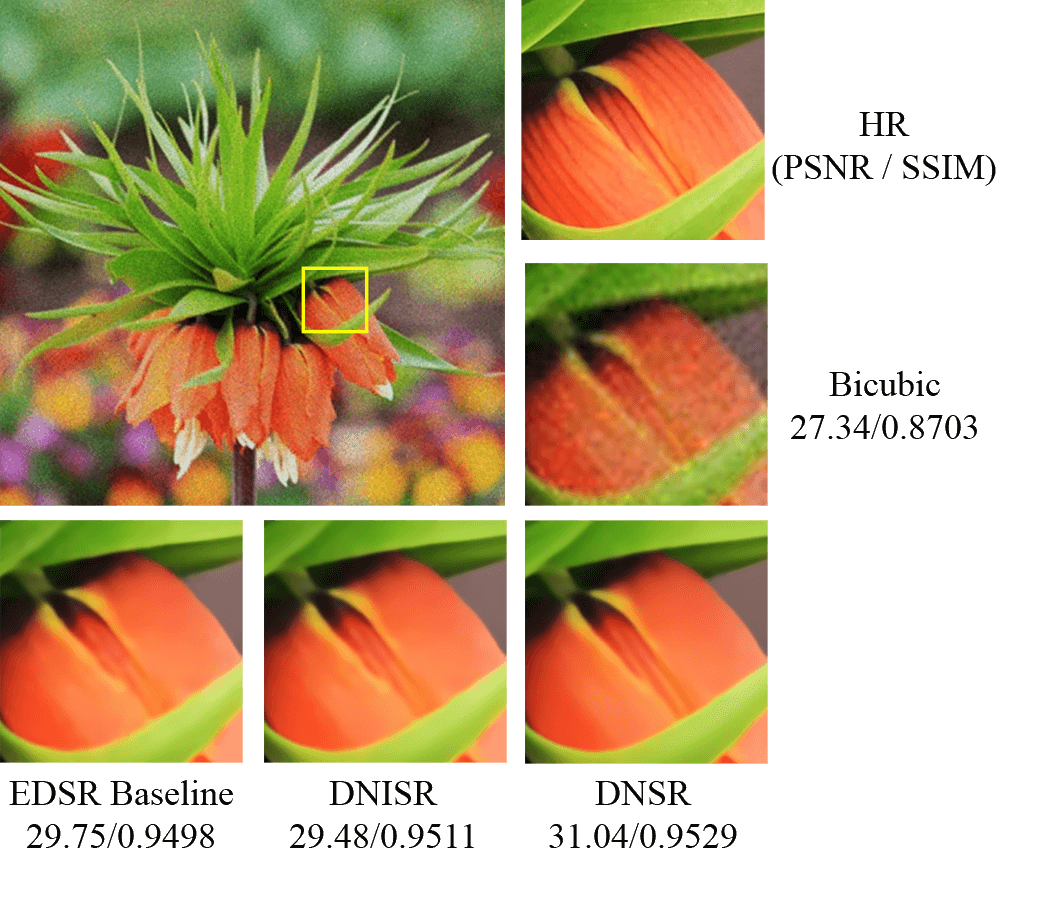}\\
    \includegraphics[width=\textwidth]{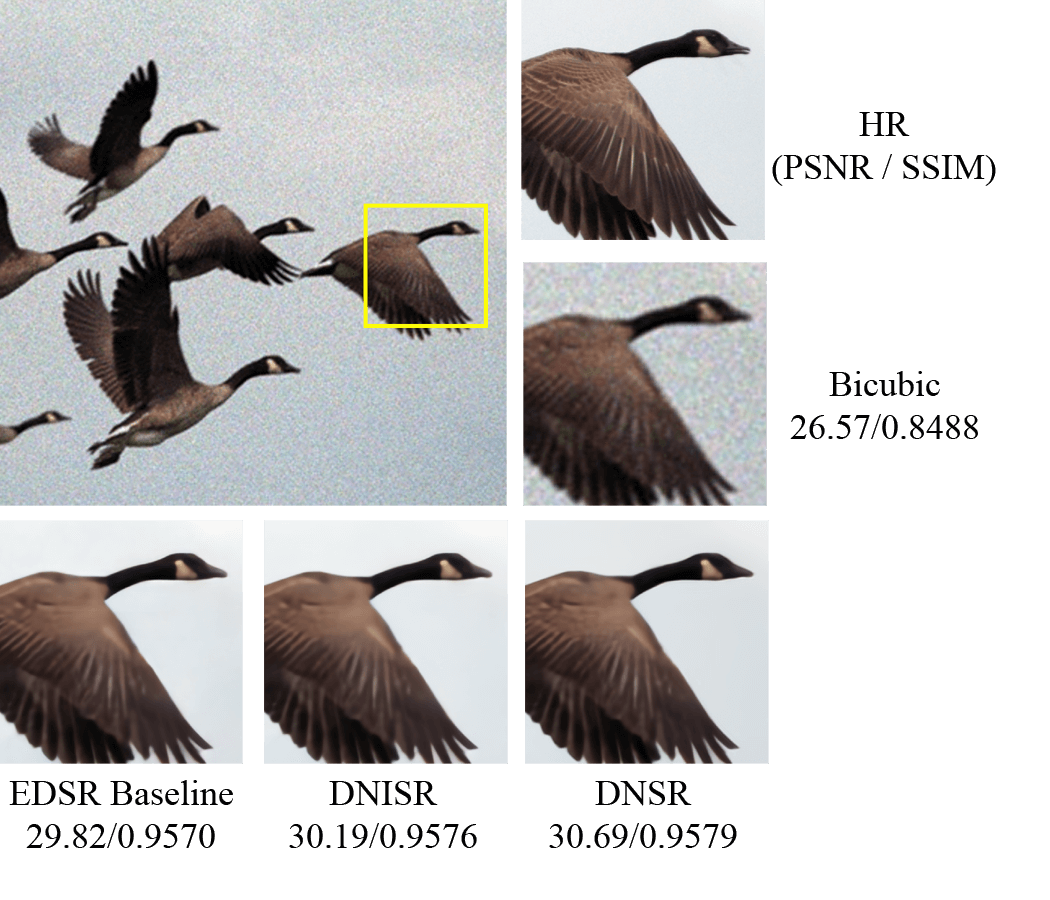}\\
    \includegraphics[width=\textwidth]{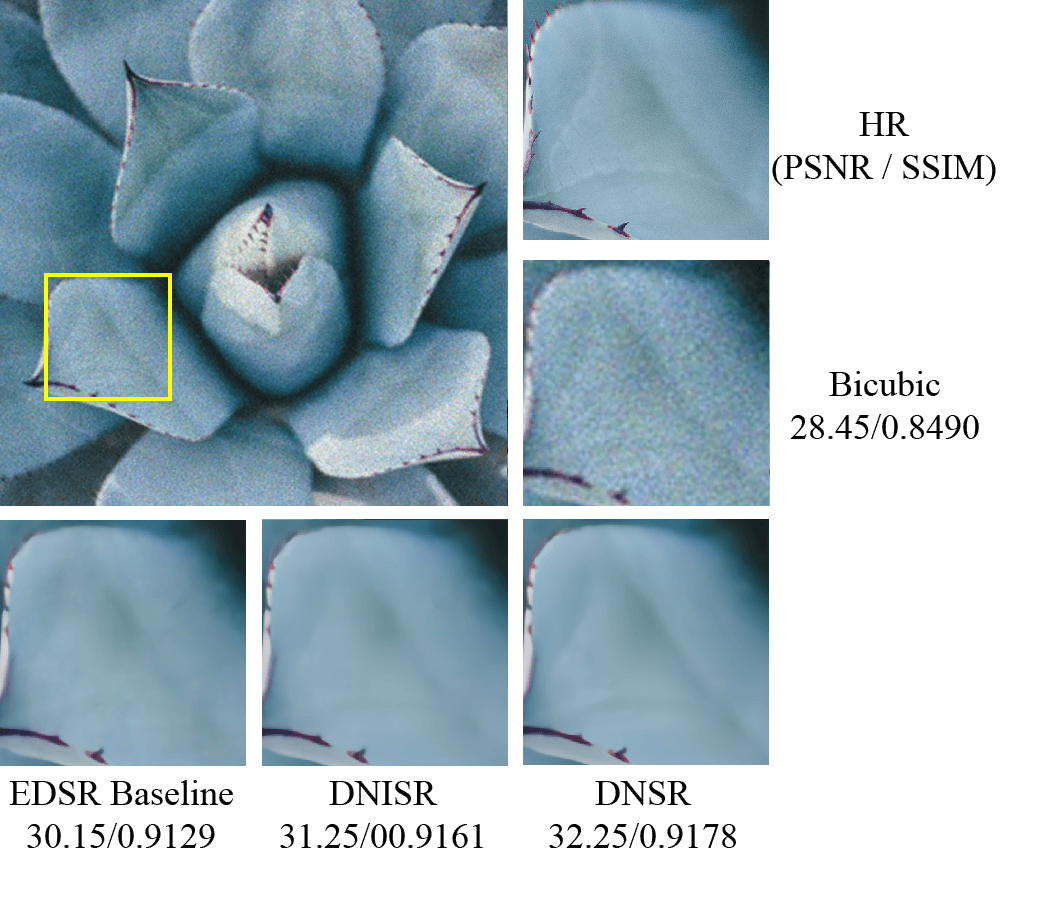}
\end{subfigure}
\caption{Comparison of different techniques for upscaling images by a factor of 8 (left) and upscaling noisy images by a factor of 4 (right). The visual differences between the images are especially pronounced in the last images of each column, where the folds in the leaves are much clearer.}
\label{fig:samples}
\end{figure*}

We discovered several tricks that can be used any time, with almost any network architecture. To see the results these tricks had on upscaling images by a factor of $8$, see Figure \ref{fig:samples}.
\begin{itemize}
	\item RGB Layer Shuffle: In addition to flipping and rotating the image patches during training and generation, we randomly shuffled the red, green, and blue layers. This improved our overall model by a small amount. This trick is applicable to any convolutional structure. Figure \ref{fig:rgbshuffling} shows the effect of test-time RGB Shuffling.
	\item Per-Image Mean Shift: Instead of calculating the average mean throughout all of the images and normalizing by that value, as in the original EDSR paper, we instead normalized each individual image patch during training by subtracting its mean.
	\item Different Upsampling Techniques: For Track 1, we started by using sub-pixel shift to upscale the image. In addition, to upsample by a factor of $8$, we concatenated three $\times 2$ upsamplers, as in the original EDSR paper. Using this approach, we ran into artifacts induced by the upscaling (see Figure \ref{fig:artifacts}). These artifacts were diminished by switching the upsampling method to Transposed Convolution upsampling. However, even with the sub-pixel shift upscaler, the problem went away when we switched to directly learning a $\times 8$ upscaler instead of three concatenated $\times 2$ upscalers.
	
	In our final method, we found that direct $\times 8$ upscaling combined with the sub-pixel shift upscaler produced images with higher PSNR values. However, the concatenated $\times 2$ upscalers seemed less prone to creating artifacts due to antialiasing (see Figure \ref{fig:antialias}).
	
	\begin{figure}[htbp]
	    \centering
	    \includegraphics[width=\columnwidth]{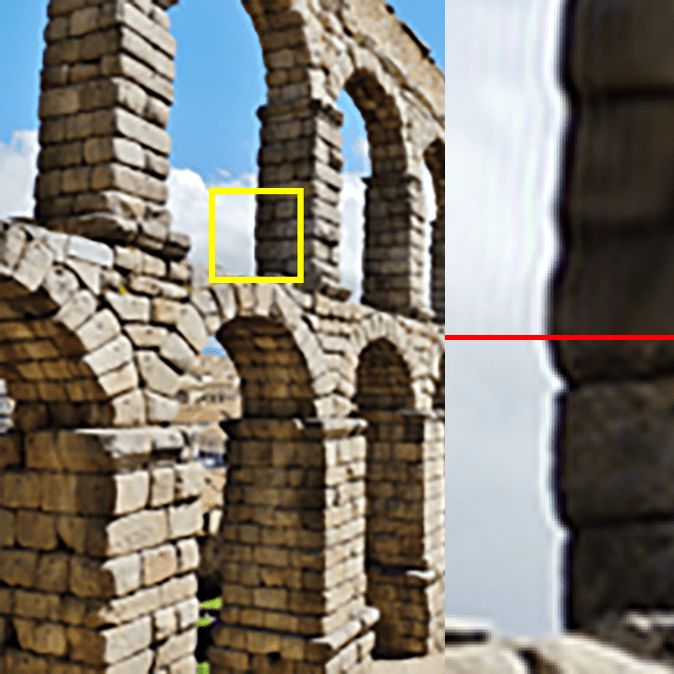}
	    \caption{The upscaler using sub-pixel shift (top-right) has clear chromatic artifacts, while the upscaler using transposed convolutional upscaling (bottom-right) does not.}
	    \label{fig:artifacts}
	\end{figure}
	
	\begin{figure}[htbp]
	    \centering
	    \includegraphics[width=\columnwidth]{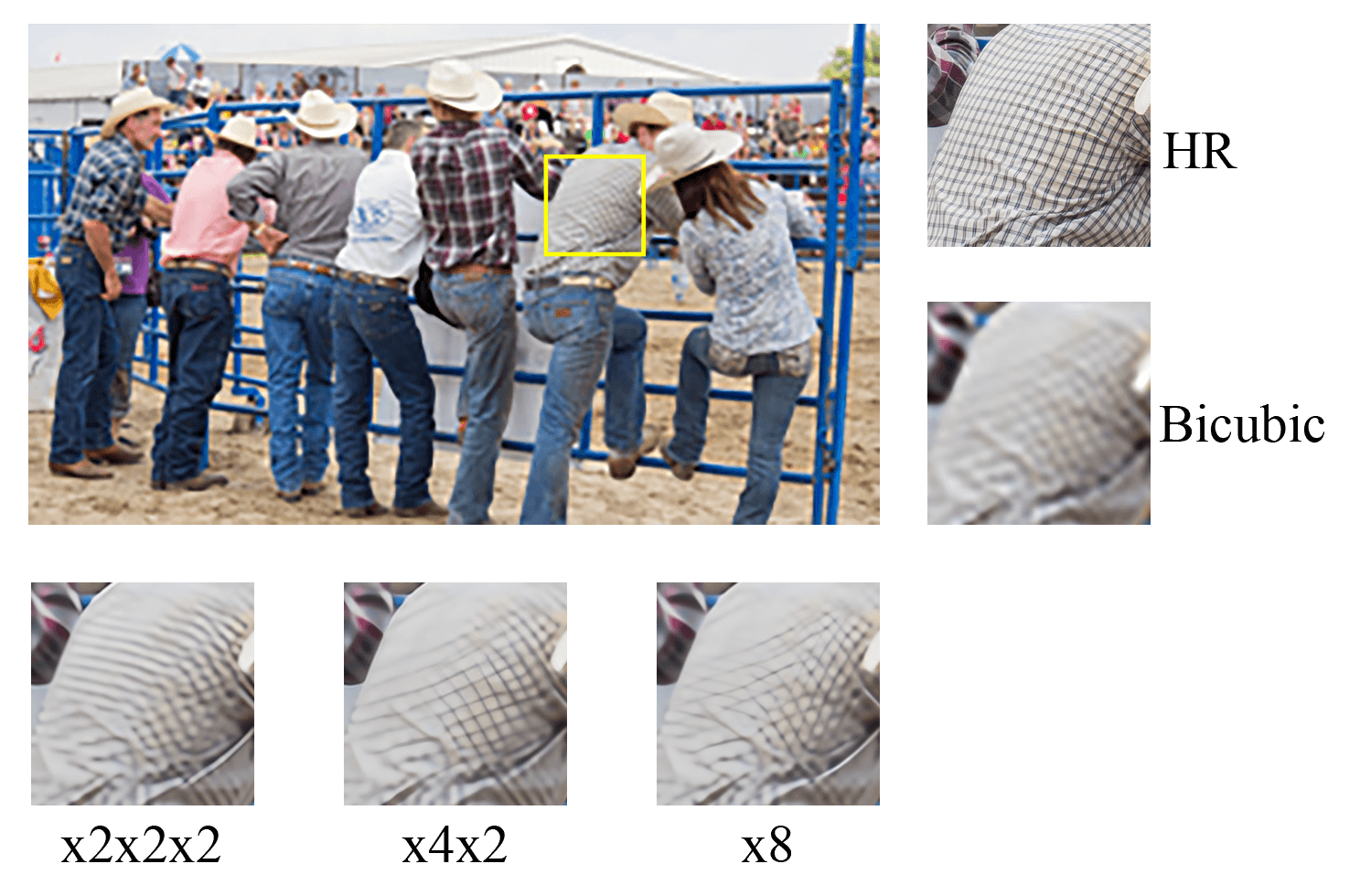}
	    \caption{Diagonal lines created by some upscaling methods due to anti-aliasing}
	    \label{fig:antialias}
	\end{figure}
	
	\item Residual Scaling Factor: In EDSR, each residual layer is multiplied by $0.1$ at the end. Instead of hardcoding this parameter, we allowed it to be a free variable that could be trained.
	\item Edge Loss: We attempted to add an edge-loss component to the loss by applying a Sobel filter to both the upscaled and ground-truth images, and comparing those. However, this did not improve on our previous model.
	\item Kernel Size: We tried various kernel sizes, however $2\times2$ produced worse results and we could not successfully train the network with the $4\times4$ and $5\times5$ kernel sizes.
\end{itemize}

\begin{figure}
    \centering
    \includegraphics[width=\columnwidth]{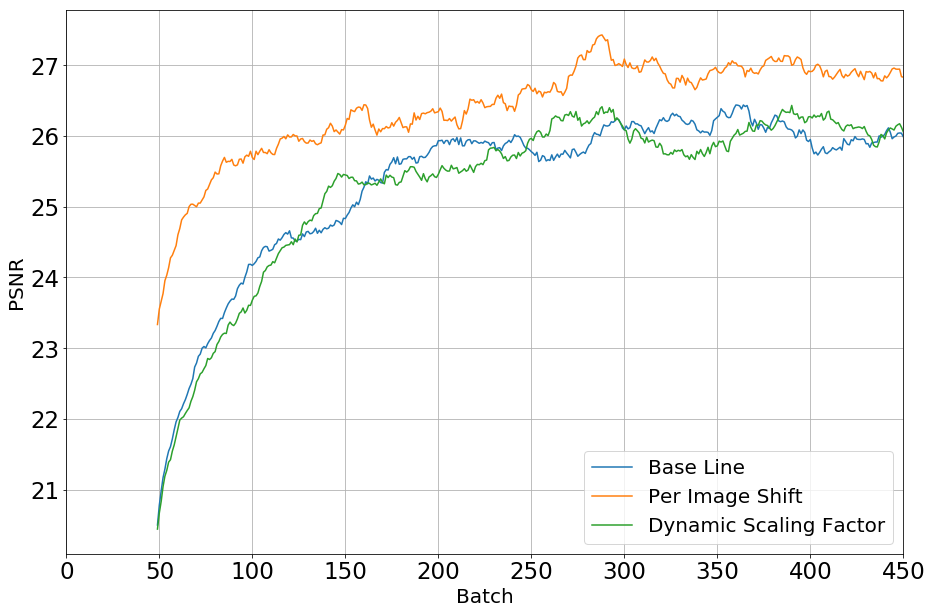}
    \caption{Convergence rates for baseline EDSR, EDSR with per image shift and dynamic scaling factor. The figure is plotted with rolling mean of 50.}
    \label{fig:convergence}
\end{figure}

Figure \ref{fig:convergence} shows a comparison of convergence rates for baseline EDSR, EDSR with per image intensity shift, and EDSR with dynamic residual scaling factors. All training started with randomly initialized weights. Using per-image mean shift gives a higher initial PSNR and faster convergence.

\begin{figure}[htbp]
\centering
\includegraphics[width=\columnwidth]{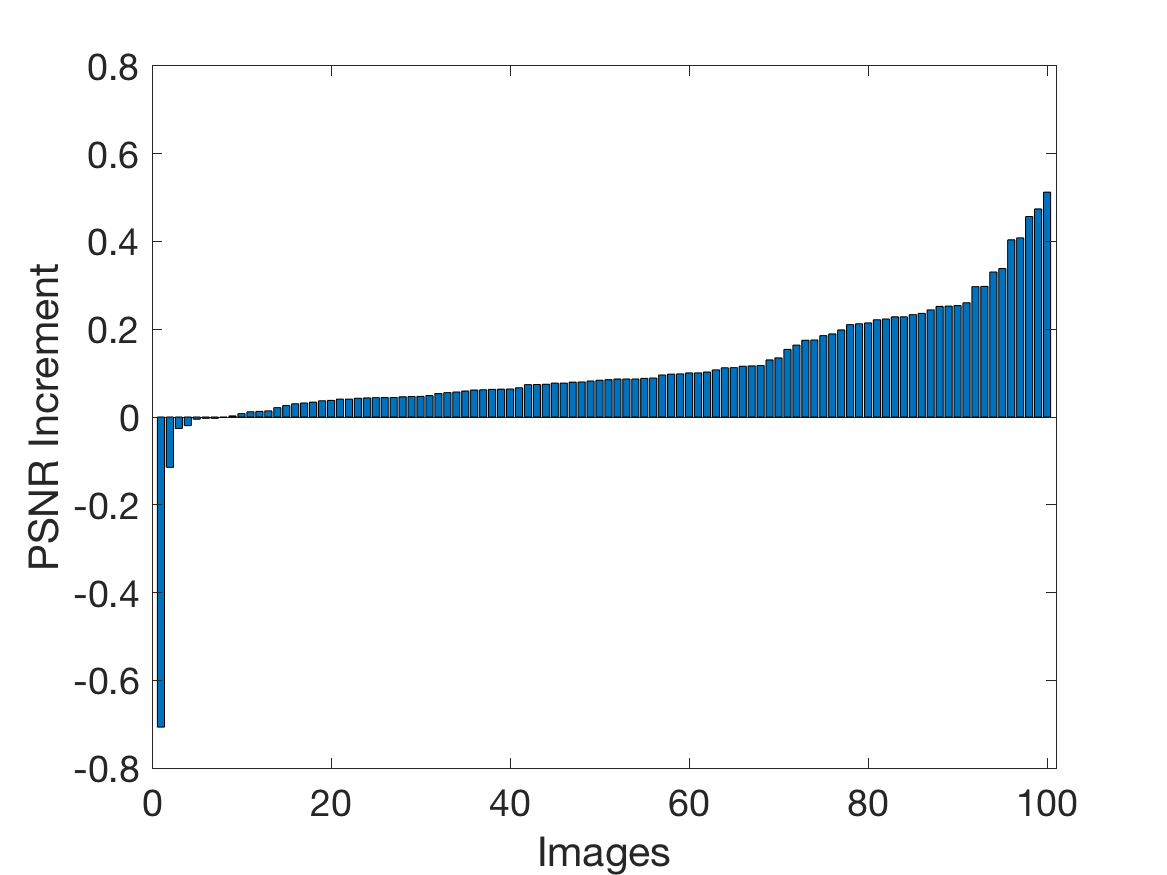}\\
\caption{PSNR difference between our improved EDSR and baseline EDSR (sorted by difference in PSNR) on the 100 image validation set from DIV2K \cite{Agustsson_2017_CVPR_Workshops}.}
\label{fig:track1diff}
\end{figure}

Table \ref{tbl:tricks} and Figure \ref{fig:track1diff} show a PSNR comparison for VDSR, EDSR, and our improved EDSR model. For a visual comparison of the images produced by each algorithm, see Figure \ref{fig:samples}.

\begin{table}[ht!]
    \centering
    \begin{tabular}{c|c|c|c|c}
       \footnotesize{Algorithm}  & \footnotesize{EDSR} & \footnotesize{Improved} \footnotesize{EDSR} & \footnotesize{VDSR} & \footnotesize{BICUBIC}\\ \hline
        \footnotesize{PNSR} & 25.49 & \textcolor{red}{25.60} & 24.70 & 23.69 \\
        \footnotesize{SSIM} & 0.6930 & \textcolor{red}{0.6974} & 0.6580 & 0.6291\\
    \end{tabular}
    \caption{Comparison of our improved EDSR algorithm to several baselines. The improvements could have been made to any SR algorithm besides EDSR. The numbers reported were computed on the DIV2K validation data set.}
    \label{tbl:tricks}
\end{table}

Figure \ref{fig:rgbshuffling} shows the PSNR increment across the 100 images from the DIV2K validation set after applying only RGB Shuffling to EDSR. In 97 out of the 100 cases, there was a boost in PSNR.
\section{Conclusion}
We discussed two new network architectures for denoising and preserving general structure in images during super-resolution, as well as a toolbox of tricks. The vast majority of the findings described here were not implemented in time for the competition deadline. Our high-scoring entries are mostly a result of the toolbox of tricks discussed above. 
We have noticed substantial improvement from our competition entries to the results reported in this paper.

Our code is available at: \url{https://github.com/websterbei/EDSR_tensorflow} and \url{https://github.com/nikhilvravi/DukeSR}.

{\small
\bibliographystyle{ieee}
\bibliography{egbib}
}

\end{document}